\documentclass[10pt,twocolumn,letterpaper]{article}

\usepackage{cvpr}              % To produce the CAMERA-READY version
\usepackage[dvipsnames]{xcolor}

\definecolor{cvprblue}{rgb}{0.21,0.49,0.74}
\usepackage[pagebackref,breaklinks,colorlinks,citecolor=cvprblue]{hyperref}

\title{VI-Diff: Unpaired Visible-Infrared Translation Diffusion Model for Single Modality Labeled  Visible-Infrared Person Re-identification}

\author{Han Huang$^{1,2,3}$   Yan Huang$^{2,3}$   Liang Wang$^{2,3}$\\
$^1$University of Chinese Academy of Sciences (UCAS)\\
$^2$The Center for Research on Intelligent Perception and Computing (CRIPAC)\\ $^3$Institute of Automation, Chinese Academy of Sciences (CASIA)\\
{\tt\small \{han.huang, yan.huang\}@cripac.ia.ac.cn, wangliang@nlpr.ia.ac.cn}
}

\begin{document}
\maketitle
\begin{abstract}
Visible-Infrared person re-identification (VI-ReID) in real-world scenarios poses a significant challenge due to the high cost of cross-modality data annotation. Different sensing cameras, such as RGB/IR cameras for good/poor lighting conditions, make it costly and error-prone to identify the same person across modalities. To overcome this, we explore the use of single-modality labeled data for the VI-ReID task, which is more cost-effective and practical. By labeling pedestrians in only one modality (e.g., visible images) and retrieving in another modality (e.g., infrared images), we aim to create a training set containing both originally labeled and modality-translated data using unpaired image-to-image translation techniques. In this paper, we propose VI-Diff, a diffusion model that effectively addresses the task of Visible-Infrared person image translation. Through comprehensive experiments, we demonstrate that VI-Diff outperforms existing diffusion and GAN models, making it a promising solution for VI-ReID with single-modality labeled data. Our approach can be a promising solution to the VI-ReID task with single-modality labeled data and serves as a good starting point for future study. Code will be available.
\end{abstract}    
\section{Introduction}
Person re-identification (ReID) has long been widely studied due to its importance in security and surveillance systems. In this field, Visible-Infrared ReID (VI-ReID) handles cross-modality identity matching between visible images and infrared images. Previous research has established numerous VI-ReID models achieving promising results in accuracy.

However, beyond focusing solely on model performance, practical challenges exist in cross-modality data annotation that have been unexplored. Constructing a training set necessitating annotations for images of the same person from both RGB and IR cameras in real-world scenarios poses difficulties and several reasons contribute to this. \textbf{First}, due to visual differences, a person's appearance in a visible image can markedly differ from their infrared image counterpart, demanding expert annotators. Despite expertise, establishing direct visual correspondence between modalities remains challenging and uncertain. \textbf{Second}, individuals might be captured by RGB cameras during the day but invisible to IR cameras at night, causing disjoint data and complicating pair matching across modalities. \textbf{Lastly}, changes in environmental conditions like lighting, weather, and surroundings can exacerbate visual gaps between visible and infrared images, introducing noise and inconsistency during annotation. These challenges hinder the collection and annotation of cross-modality data, resulting in a labor-intensive, time-consuming, and error-prone process.

\begin{figure}
    \centering
    \begin{subfigure}{0.4\linewidth}
        \includegraphics[width=1\linewidth]{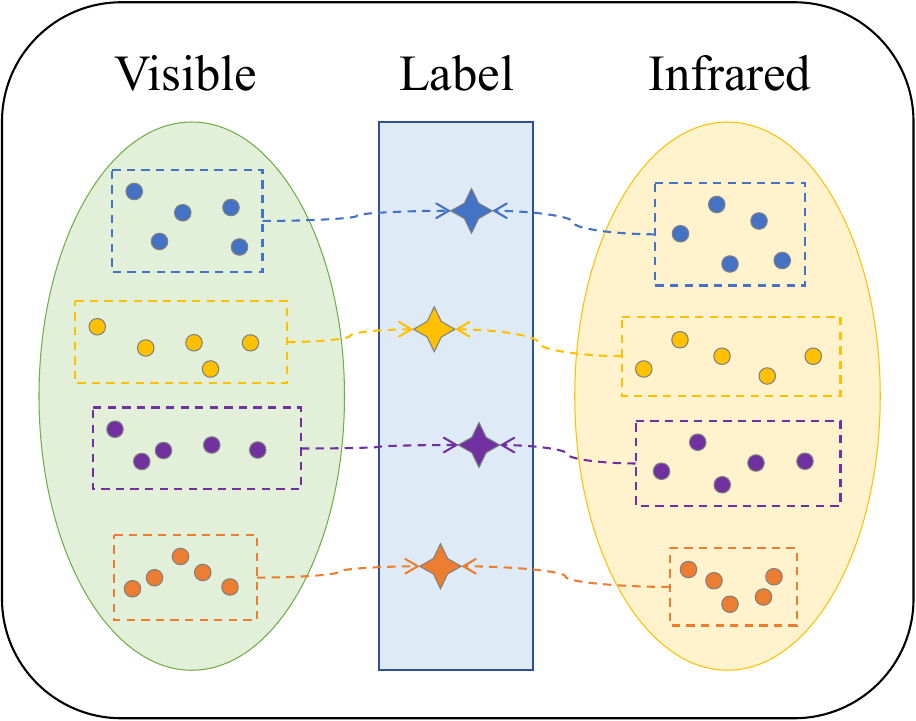}
        \caption{Supervised VI-ReID}
        \label{supervised}
    \end{subfigure}
    \begin{subfigure}{0.4\linewidth}
        \includegraphics[width=1\linewidth]{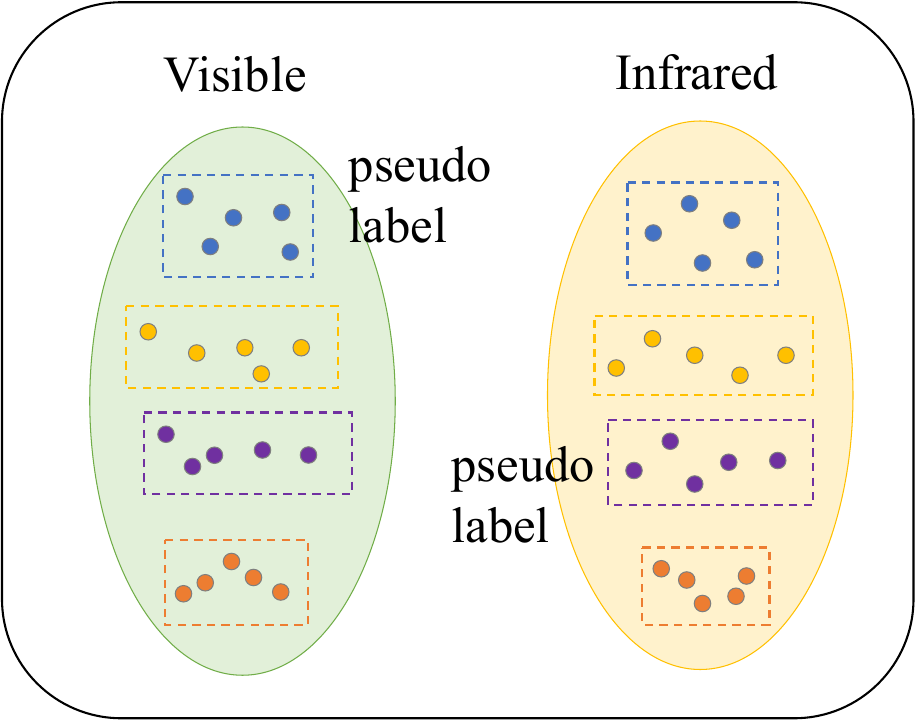}
        \caption{Unsupervised VI-ReID}
        \label{unsupervised-pseudo}
    \end{subfigure}
    \begin{subfigure}{0.4\linewidth}
        \includegraphics[width=1\linewidth]{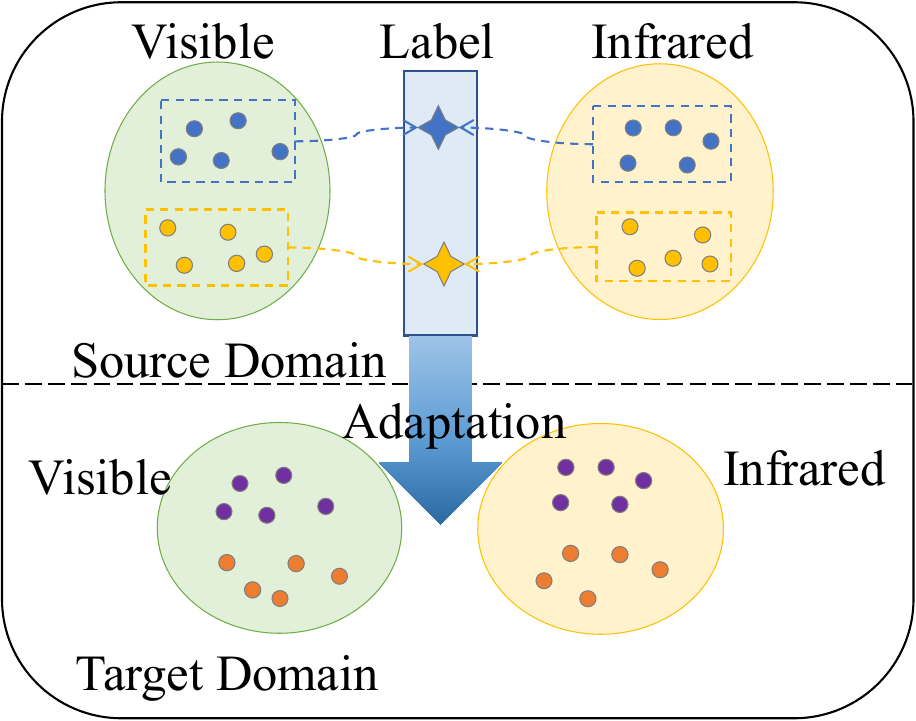}
        \caption{Unsupervised VI-ReID with Domain Adaptation}
        \label{unsupervised-adaptation}
    \end{subfigure}
    \begin{subfigure}{0.4\linewidth}
        \includegraphics[width=1\linewidth]{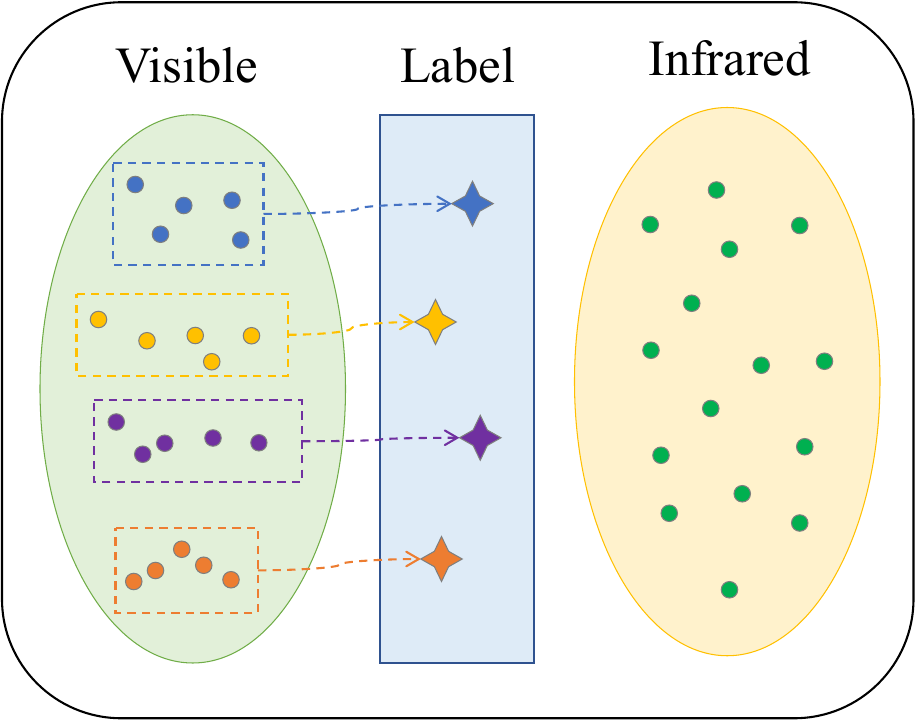}
        \caption{Single Modality Labeled VI-ReID (our setting)}
        \label{single-labled}
    \end{subfigure}
    \caption{Different VI-ReID settings. (a): Supervised VI-ReID; (b): Unsupervised VI-ReID, training with pseudo label; (c): Unsupervised VI-ReID with extra labeled data and domain adaptation; (d): Single modality labeled VI-ReID, data from one modality has unknown identities.}
    \label{fig:VI-ReID-class}
\end{figure}

Given these concerns, a compelling subject for investigation emerges: Can we leverage single-modality labeled data to address the challenging cross-modality VI-ReID task? The approach involves dealing with entirely unknown infrared (visible) images by focusing on labeling only visible (infrared) images, which is a more achievable task. In doing so, we attempt VI-ReID even in the absence of labels from one modality. This innovative setting aims to alleviate the heavy burdens associated with cross-modality data annotation, with the key lying in effectively utilizing labeled data from a single modality. By exploring this strategy, we may potentially overcome the limitations posed by cross-modality data annotation, making the VI-ReID task more feasible and practical in real-world scenarios.

In Fig.\ref{fig:VI-ReID-class}, we emphasize the distinction between fully-supervised VI-ReID, unsupervised VI-ReID \cite{LiangH2H}, and VI-ReID with single modality labeled data (our setting). While fully-supervised VI-ReID is well-studied, limited work has tackled unsupervised VI-ReID, which suffers from lacking annotations for both modalities and resulting in severe performance decline. As a trade-off between fully-supervised and unsupervised VI-ReID, our task represents a novel challenge for researchers. The particular challenge of single-modality labeled VI-ReID lies in the impossibility of directly acquiring visible-infrared image pairs to train a VI-ReID model. Therefore, the focus becomes how to effectively utilize labeled data from one modality with unlabeled data from another. This unique setting opens up new possibilities for developing innovative approaches for VI-ReID, making it a promising avenue for future research.

Intuitively, to train a VI-ReID model for this task, we can employ an unpaired image-to-image translation model to translate labeled images into another modality and assign the same labels to the generated images \cite{wei2018person, deng2018image, huang2019sbsgan}. GAN-based methods have been extensively developed in this field, making them applicable to visible-infrared person image translation. Meanwhile, recent advancements in diffusion models have demonstrated superior results in generative tasks \cite{dhariwal2021diffusion}. However, despite these promising results, diffusion models have not been thoroughly investigated or directly applied to visible-infrared person image translation. To leverage the capabilities of diffusion models and address the gap in visible-infrared person image translation, we introduce VI-Diff, a diffusion model trained on unpaired visible-infrared images. By training on unpaired data, VI-Diff learns to capture the underlying distribution of both modalities, allowing it to generate visually realistic and identity-preserving images across modalities without the need for paired data or annotations.

Inspired by the recent diffusion model in image-to-image generation \cite{saharia2022palette}, our VI-Diff introduces a modality-insensitive ID-relevant condition to ensure high-quality image translation between modalities while effectively preserving the identity information compared to existing unpaired image-to-image diffusion models \cite{Choi2021ILVRCM, zhao2022egsde}. Additionally, the VI-Diff introduces a modality indicator as classifier-free guidance \cite{ho2022classifier} to ensure the feasibility and efficacy of cross-modality translation. By combining the strengths of diffusion models with a carefully designed translation approach, our VI-Diff holds great promise in advancing the state-of-the-art in visible-infrared person image translation.

\begin{figure*}
    \centering
    \includegraphics[width=0.75\linewidth]{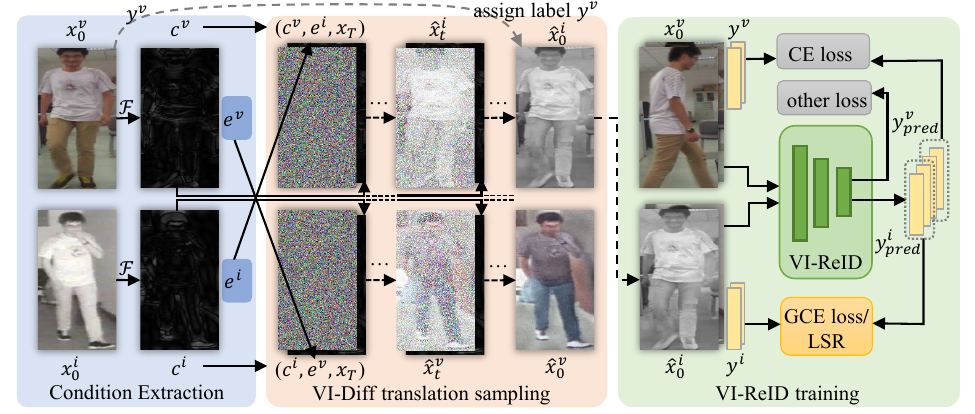}
    \caption{Overview of VI-Diff translation sampling and VI-ReID training: In VI-Diff translation, the modality indicator $e$ is altered; $x_T$ represents the random starting Gaussian noise. In VI-ReID training, the loss for the predicted identities of generated images is computed using either GCE or LSR. Refer to the \textit{Method} section for more details.}
    \label{fig:framework}
\end{figure*}

Images generated by our VI-Diff with the modality-insensitive ID-relavant condition hold the potential for maintaining the original ID labels on the generated images during inference. These generated images are combined with real images from another modality to form a dataset for VI-ReID models. However, it is possible that VI-Diff might generate samples that do not precisely match the ground-truth label of its counterpart from the other modality, resulting in noise labels that could mislead the VI-ReID model. To address this concern, we employ noise-robust loss functions such as Generalized Cross Entropy (GCE) \cite{zhang2018generalized} and Label Smoothing Regularization (LSR) \cite{szegedy2016rethinking} on those cross-modality generated images. This approach further enhances the accuracy and reliability of the VI-ReID model when utilizing cross-modality generated images for identity matching. The overview is illustrated in in Fig.\ref{fig:framework}

Our main contributions can be summarized as follows: 

1) We raise concerns regarding VI-ReID in real-world scenarios and delve into the exploration of a new task: leveraging single-modality labeled data to accomplish the cross-modality VI-ReID task.

2) We introduce VI-Diff, a diffusion model capable of visible-infrared person image translation, which contributes to good performance in VI-ReID with single-modality labeled data. To the best of our knowledge, this is the first VI-ReID work using the diffusion model.

3) We conduct comprehensive experiments to validate the effectiveness and potential of VI-Diff in addressing the challenges of cross-modality VI-ReID and advancing the state-of-the-art in this field.
\section{Related Works}
\subsection{Supervised VI-ReID}
Numerous approaches have aimed to address the cross-modality image retrieval problem by learning shared and modality-specific features. Classic models like \cite{ye2019bi} and \cite{dai2018cross} incorporate shared embeddings and generative adversarial training, respectively. Hi-CMD \cite{choi2020hi} employs pose- and illumination-invariant features for cross-modality matching through ID-preserving image generation.

Beyond modality-shared features, methods like cm-SSFT \cite{lu2020cross}, DDAG \cite{ye2020dynamic}, and LbA \cite{park2021learning} consider modality-specific traits and contextual information. Moreover, MPANet \cite{WU2021mpanet} has proposed unified frameworks for nuances discovery, and DEEN \cite{zhang2023diverse} generates embeddings, both aimed for modality discrepancy reduction.

While these techniques prioritize enhanced VI-ReID accuracy, they rely on well-annotated data from both modalities. In contrast, our work tackles the challenge of single-modality labeled data and incorporates these VI-ReID models in our experiments by leveraging our generated data with assigned pseudo labels.

%%%%%%%%%%%%%%%%%%%%%%%%%%%%%%%%%%%%%%%%%%%%%%%%%%%%%
\subsection{Unsupervised VI-ReID}
Unsupervised VI-ReID aims to alleviate the labor-intensive process of labeling person identities for both visible and infrared modalities, as many datasets lack identity annotations for both modalities. However, compared to fully-supervised VI-ReID, the number of studies tackling this problem remains limited. \cite{liang2021homogeneous} proposed a two-stage approach. The first stage generates modality-specific features and pseudo-labels, while the second stage distills shared knowledge from pseudo-labeled data, yielding modality-invariant representations. \cite{wang2022optimal} employed unsupervised domain adaptation to generate pseudo labels for the visible modality, transferring them to the infrared modality using optimal-transport strategies. Furthermore, \cite{Fu2022c2} introduced the dual alignment network to address visible-infrared cross-modal cross-domain ReID challenges.

These approaches leverage labeled datasets from external sources to generate pseudo labels for visible and infrared images, assuming corresponding identities across modalities. In contrast, our method doesn't rely on external labeled data and assumes that identities of unlabeled data are entirely unknown during training. As a result, we transfer modality information from labeled domain data by generating cross-modality labeled images, alleviating the lack of unlabeled data in one modality.

%%%%%%%%%%%%%%%%%%%%%%%%%%%%%%%%%%%%%%%%%%%%%%%%%%%%%%%
\subsection{Unpaired Image-to-Image Translation}
Unpaired image-to-image translation focuses on transforming images between different domains without requiring paired training data. Cycle-consistency is a prevalent strategy in GAN-based methods, exemplified by CycleGAN \cite{zhu2017unpaired}, DualGAN \cite{yi2017dualgan}, and DiscoGAN \cite{kim2017learning}. In these, translating an image to another domain and then back ideally reconstructs the original image. Methods like DRIT \cite{DRIT} have improved this approach by achieving diverse image translation through disentangled representations. Moreover, alternatives to cycle-consistency for one-sided mapping are proposed, such as DistanceGAN \cite{benaim2017one}, which maintains intra-domain image distances, and CUT \cite{Park2020CUT}, which enhances mutual information between inputs and generated images through contrastive learning. Multi-domain models like StarGANv2 \cite{choi2020stargan} bring about better architectures and regularization techniques, enabling more adaptable and stable cross-domain image generation.

Recent investigations have integrated diffusion models into image-to-image translation. DDIB \cite{su2022dual} employs two independently trained models for the source and target domains. Translation involves deriving latent noise encoding from the source image using the source diffusion model and then applying the reverse diffusion process to reconstruct the target image with the target diffusion model. However, this approach lacks a guarantee of preserving person identity and detail in visible-infrared image translation, with modality gaps leading to substantial latent noise differences. On the other hand, ILVR \cite{Choi2021ILVRCM} employs a diffusion model solely in the target domain and refines sampling using the source image. EGSDE \cite{zhao2022egsde} introduces an energy function based on feature extractors trained on both domains to encourage preserving domain-independent features in translated images. Nevertheless, these methods, lacking specific modality gap attention, yield suboptimal guidance for visible-infrared person translation, resulting in significantly inferior translation performance when contrasted with our VI-Diff model.

\section{Method}

Before introducing our method, we first revisit the classic Denoising Diffusion Probabilistic Models (DDPM) \cite{ho2020denoising}. During training, DDPM gradually adds Gaussian noise to an input image $x_0$ to create a sequence of noised images $x_1, x_2, \ldots, x_T$. At each timestep, the noise level increases until the final noised image $x_T$ becomes a standard Gaussian distribution $x_T \sim \mathcal{N}(0, \mathrm{I})$. The model then learns to reverse this noising process to transform the standard Gaussian back to the original image $x_0$.

The noised image $x_{t\in[1,T]}$ is sampled from the distribution $q(x_t| x_0)$, where a series of $\alpha_t \in (0,1)$, and $\bar\alpha_t := \prod_{i=1}^t \alpha_i$ are pre-set parameters. The random Gaussian noise $\epsilon$ is also involved in the sampling process:
\begin{align}
q(x_t| x_0) &= \mathcal{N}(x_t; \sqrt{\bar\alpha_t}x_0, (1-\bar\alpha_t)\mathbf{I}) \label{qtheta} \\
&= \sqrt{\bar\alpha_t}x_0 + \epsilon\sqrt{1-\bar\alpha_t}, \quad \epsilon \sim \mathcal{N}(0, \mathrm{I}). \label{eps}
\end{align}

Here, $q(x_t| x_0)$ represents the Gaussian distribution with mean $\sqrt{\bar\alpha_t}x_0$ and covariance matrix $(1-\bar\alpha_t)\mathbf{I}$. The parameter $\bar\alpha_t$ is the cumulative product of $\alpha_i$ for $i$ from $1$ to $t$, and it controls the amount of noise added at each timestep during the diffusion process.

To generate a slightly denoised image $x_{t-1}$ from $x_t$, we can sample $x_{t-1}$ from the distribution $p_\theta(x_{t-1}| x_t)$:
\begin{align}
p_\theta(x_{t-1}| x_t) &:= \mathcal{N}(x_{t-1}; \mu_\theta(x_t,t), \Sigma_\theta(x_t,t)). \label{ptheta}
\end{align}

Here, $\theta$ represents the parameters of the neural network, which is typically a variant of UNet \cite{ronneberger2015u}. The neural network is trained to predict the mean $\mu_\theta(x_t,t)$ and the diagonal covariance matrix $\Sigma_\theta(x_t,t)$ for the distribution given in Equ.\ref{qtheta}.

The predicted mean $\mu_\theta(x_t,t)$ and covariance matrix $\Sigma_\theta(x_t,t)$ are used to construct the Gaussian distribution $p_\theta(x_{t-1}| x_t)$ from which the denoised image $x_{t-1}$ is sampled. By utilizing this denoising process iteratively for each timestep during the reverse process, the model can reconstruct the original image $x_0$ from the final noised image $x_T$.

DDPM does not directly predict $\mu_\theta(x_t,t)$ and $\Sigma_\theta(x_t,t)$. Instead, it predicts $\epsilon_\theta(x_t,t)$ for $\epsilon$ in Equ.\ref{eps}, which contributes to improved sample quality. This approach simplifies the objective by directly comparing the difference between predicted noise and real noise added to the image:
\begin{align}
L_{simple} := E_{t, x_0, \epsilon}\left[ \parallel \epsilon-\epsilon_\theta(x_t, t) \parallel^2 \right], \label{equ:L_simple}
\end{align}
and $\mu_\theta(x_t,t)$ is calculated from $\epsilon_\theta(x_t,t)$ as follows:
\begin{align}
\mu_\theta(x_t,t) &= \frac{1}{\sqrt{\alpha}}\left(x_t - \frac{1-\alpha_t}{\sqrt{1-\bar\alpha_t}}\epsilon_\theta(x_t,t)\right). \label{mu_theta}
\end{align}

For enhanced sample quality, $\Sigma_\theta(x_t,t)$ is fixed to a constant $\Sigma_\theta$ in DDPM. With $\mu_\theta(x_t,t)$ and $\Sigma_\theta$, a slightly denoised image $x_{t-1}$ is sampled as:
\begin{align}
x_{t-1} &= \mu_\theta(x_t,t) + \Sigma_\theta\epsilon, \quad \epsilon \sim \mathcal{N}(0,\mathrm{I}). \label{x_t-1_sample}
\end{align}

Lastly, the fully denoised image $x_0$ is generated from $x_T$ through iterative denoising timesteps from $T$ to $0$.
%%%%%%%%%%%%%%%%%%%%%%%%%%%%%%%%%%%%%%%%%%%%%%%%%%%%%%%%%%%%%%%%%%%%%%%%%%
\subsection{Modality-Insensitive ID-Relevant Condition}
\begin{figure}
    \centering
    \includegraphics[width=0.4\textwidth]{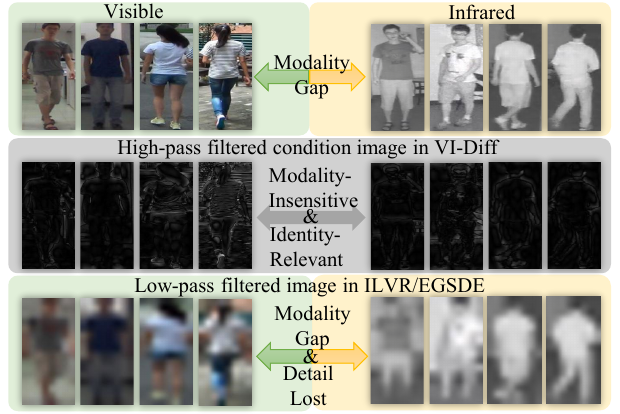}
    \caption{Modality-insensitive ID-relevant condition for VI-Diff, in comparison with the low-pass filtered image.}
    \label{fig:highpass}
\end{figure}

DDPM is an unconditional diffusion model that lacks control over the contents of sample results. To achieve visible-infrared image translation with preserved person details, we need to find a modality-insensitive ID-relevant condition ($c$) for our VI-Diff model.

Given the condition $c$, the diffusion model accepts $c$ as input with $x_t$ and $t$, resulting in modifications to Equ.\ref{equ:L_simple} and Equ.\ref{mu_theta}:
\begin{align}
L_{simple}:=E_{t,x_0,c\sim p(x_0,c),\epsilon}\left[ \parallel \epsilon-\epsilon_\theta(x_t,t,c) \parallel^2 \right], \\
\mu_\theta(x_t, t, c) = \frac{1}{\sqrt{\alpha_t}}(x_t- \frac{1-\alpha_t}{\sqrt{1-\bar\alpha_t}}\epsilon_\theta(x_t,t,c)).
\end{align}

We use the high-pass filtered pedestrian image as the condition $c$. Given visible and infrared images denoted as $x_0^v$ and $x_0^i$, we apply the high-pass filter $\mathcal{F}$ to obtain conditions $c^v=\mathcal{F}(x_0^v)$ and $c^i=\mathcal{F}(x_0^i)$. Both $c^v$ and $c^i$, passed through the high-pass filter, are insensitive to modality information and preserve ID-related features by retaining modality-insensitive high-frequency details (e.g., edges and textures of the image, as shown in Fig.\ref{fig:highpass}).

By training VI-Diff on datasets containing both visible and infrared images, represented as $\mathcal{D}$=$\{\left(x_0^v, c^v \right), \left(x_0^i, c^i \right) \}$, we achieve better visible-infrared image translation compared to unconditional diffusion models used in ILVR \cite{Choi2021ILVRCM} and EGSDE \cite{zhao2022egsde}, which lose image details and disregard the modality gap between visible and infrared images by refining the sampling process with low-pass filtered source images.

%%%%%%%%%%%%%%%%%%%%%%%%%%%%%%%%%%%%%%%%%%%%%%%%%%%%%%%
\subsection{Modality Indicator and Guidance}
The conditions $c^v$ and $c^i$ are effective in controlling sample contents, but they do not provide information about the modality of the generated results. Therefore, we embed modality information into the diffusion model using modality indicators $e^v=0$ and $e^i=1$ to guide the generation of visible and infrared modalities, respectively. The indicators $e$ is efficiently incorporated into the intermediate layers of the diffusion models through embedding layers, enabling \textit{classifier-free guidance} \cite{ho2022classifier}. The noise $\epsilon_\theta$ is then computed using a linear combination:
\begin{align}
    \epsilon_\theta = (1+\omega) \epsilon_\theta(x_t,t,e) - \omega\epsilon_\theta(x_t,t),
\end{align}
where $\omega$ is a weight to balance the conditional ($\epsilon_\theta(x_t,t,e)$) and unconditional ($\epsilon_\theta(x_t,t)$) noise estimations.

During training, we use $e$ that corresponds to the modality of input images together with condition $c$. In the translation sampling process, we alter $e$ at each timestep as follows:
\begin{align}
    \mathrm{training}:\ \epsilon_\theta(x_t^v,t, c^v, e^v), \epsilon_\theta(x_t^i,t, c^i, e^i),\\
    \mathrm{sampling}:\ \epsilon_\theta(x_t^v,t, c^v, e^i), \epsilon_\theta(x_t^i,t, c^i, e^v).
\end{align}

Compared with the \textit{classifier guidance} introduced by \cite{dhariwal2021diffusion}, \textit{classifier-free guidance} eliminates the need for training an extra classifier model $p_\theta(y|x_{t-1})$, which would require class labels $y\in$ (visible or infrared). While \textit{classifier guidance} is useful for guiding pre-trained diffusion models to generate samples belonging to different classes, it comes with an additional cost. However, in our case, we do not have a pre-trained visible-infrared diffusion model available, making training an extra modality classifier model impractical.

%%%%%%%%%%%%%%%%%%%%%%%%%%%%%%%%%%%%%%%%%%%%%%%%%%%%%%%%%%%%
\subsection{Noisy Label Smoothing}\label{sec:noisylabel}
In our single-modality labeled VI-ReID task, the datasets consist of $\{(x_0^{v_j},y^{v_j}), (x_0^{i_k}, \phi )\}$, $j$=${1, 2,\ldots,N^v}$, $k$=${1,2,\ldots,N^i}$, where $y$ is the identity label, $\phi$ is the missing label, and $N^v$ and $N^i$ are the numbers of visible and infrared images, respectively. After translation, the datasets used for VI-ReID model training consist of $\{(x_0^{v_j}, y^{v_j}), (\hat{x}_0^{i_j}, y^{v_j})\}$, where the image $\hat{x}_0^{i_j}$ is translated from $x_0^{v_j}$ and assigned the same label $y^{v_j}$.

However, the translation process might inaccurately align generated images with assigned labels, which can mislead the VI-ReID model training. To address this issue, we adopt Generalized Cross Entropy (GCE) loss \cite{zhang2018generalized} or label smoothing \cite{szegedy2016rethinking} to mitigate the impact of noisy labels. The GCE loss is formulated as:
\begin{align}
    L_q(f(x), y_j) = \frac{1-f_j(x)^q}{q},
\end{align}
where $f(\cdot)$ is the classifier, $y_j$ is the label of $x$, and the weight $q\in(0,1]$. The GCE loss is equivalent to categorical Cross Entropy (CE) for $\lim_{q\to 0}L_q$ and becomes mean absolute error (MAE) loss when $q$=$1$, which is known to be noise-robust \cite{zhang2018generalized}. For label-smoothing regularization (LSR), the one-hot label vector $y$ is replaced by smoothed vector $\hat{y}$:
\begin{align}
    \hat{y}=(1-\alpha)y+\frac{\alpha}{K},
\end{align}
where $\alpha$ is a weight, $K$ is the number of identities, and $\hat{y}$ encourages the model to be less confident in its predictions.

We apply GCE loss and LSR in VI-ReID models on generated data $\{\left(\hat{x}_0^{i_j}, y^{v_j} \right)\}_{j=1}^{N^v}$ as a substitute for identity CE loss, while real data $\{\left(x_0^{v_j}, y^{v_j} \right)\}_{j=1}^{N^v}$ continues to use CE loss. This approach helps mitigate the issue of noisy labels caused by the modality translation process and leads to improved performance in off-the-shelf VI-ReID models when those cross-modality generated samples (e.g., from visible images to infrared images + assigned $y^{v_j}$) are trained together with real samples (visible images + ground-truth $y^{v_j}$).
\section{Experiments}
\textbf{Datasets:}
In our experiments, we employ the SYSU-MM01 \cite{wu2017rgb} dataset, which is widely recognized as a benchmark for VI-ReID tasks. This dataset encompasses 491 individuals captured from six distinct non-overlapping view cameras, consisting of four visible cameras and two infrared cameras. The training dataset consists of 22,258 visible images and 11,909 infrared images. The testing dataset comprises 3,803 infrared images for query purposes and 301 visible images for the gallery. \textbf{For our specific single-modality labeled setting}, we omit the labels of one modality from original training set. If not specifically indicated, we assume that the images without labels are infrared images, as it is easier to label visible images compared to infrared ones. Thus, the real visible images and generated infrared images are used to train VI-ReID models. 

\begin{table*}
\small
\centering
\begin{tabular}{lccccclccccc}
\hline
{Model} & \multicolumn{5}{c}{All-search}                                                     &           & \multicolumn{5}{c}{Indoor search}                                                  \\ \cline{2-6} \cline{8-12} 
                       & R1             & R5             & R10            & R20            & mAP            &           & R1             & R5             & R10            & R20            & mAP            \\ \hline
CycleGAN~\cite{zhu2017unpaired}               & 15.98          & 38.04          & 51.94          & 67.63          & 18.10          &           & 19.73          & 42.92          & 58.56          & 76.89          & 27.10          \\
DRIT~\cite{DRIT}                   & 10.47          & 26.06          & 37.76          & 52.93          & 10.68          &           & 12.86          & 33.51          & 48.93          & 68.13          & 20.38          \\
StarGANv2~\cite{choi2020stargan}             & 18.54          & 41.47          & 54.04          & 69.02          & 19.78          &           & 21.15          & 46.93          & 61.36          & 78.63          & 30.29          \\
CUT~\cite{Park2020CUT}                    & 22.24          & 46.00          & 58.81          & 72.80          & 23.81          &           & 27.27          & 56.53          & 71.16          & 85.53          & 37.40          \\ \hline
Uncond-Diff~\cite{ho2020denoising}                   & 1.58           & 6.24           & 11.77          & 22.52          & 3.86           &           & 3.07           & 10.38          & 19.27          & 37.27          & 8.33           \\
ILVR~\cite{Choi2021ILVRCM}                   & 3.70           & 12.37          & 20.45          & 34.11          & 6.24           &           & 5.61           & 19.21          & 30.99          & 50.32          & 12.35          \\
EGSDE~\cite{zhao2022egsde}                  & 9.62           & 25.42          & 37.70          & 54.04          & 12.19          &           & 14.65          & 37.84          & 52.71          & 70.84          & 23.84          \\ \hline
VI-Diff (Ours)                & \textbf{24.69} & \textbf{51.11} & \textbf{64.19} & \textbf{77.79} & \textbf{24.97} & \textbf{} & \textbf{30.88} & \textbf{61.16} & \textbf{74.85} & \textbf{87.71} & \textbf{40.54} \\ \hline
\end{tabular}
\caption{Comparison of GANs (first part) and diffusion models (second part) on SYSU-MM01, results from AGW. R1, 5, 10, 20 denote Rank-1, 5, 10, 20 accuracy (\%). Uncond-Diff is an unconditional diffusion model trained on infrared images.}
\label{tab:models}
\end{table*}

\textbf{Implementations:}
We use a variant of UNet introduced by~\cite{dhariwal2021diffusion}, which has residual blocks and attention heads. The condition image $c$ is concatenated with the input image. The embeddings of modality indicator $e$ are added into each intermediate block. We adapt the network to accept input image of size $128\times 256$. The total timestep $T$ is set to $1000$, and $\alpha_1$=$1-10^{-4}$ decreases linearly to $\alpha_T$=$1-2\times10^{-2}$. We train the model on two NVIDIA Tesla V100 GPUs for 200k steps with batchsize set to 8. The AdamW optimizer is employed and the learning rate is set to $10^{-4}$. We use DDIM \cite{song2020denoising} to accelerate sampling with 25 timesteps.

\subsection{Comparison with GANs And Diffusion Models}
We begin by comparing VI-Diff with other unpaired image-to-image GANs and diffusion models, including CycleGAN \cite{zhu2017unpaired}, DRIT \cite{DRIT}, StarGANv2 \cite{choi2020stargan}, CUT \cite{Park2020CUT}, ILVR \cite{Choi2021ILVRCM}, and EGSDE \cite{zhao2022egsde}. These models are trained using publicly available source code. Additionally, for comparison, we also evaluate an unconditional diffusion model \cite{ho2020denoising} on infrared images. In this model, the generated images are assigned ID labels that each dimension has the same probability value of $\frac{1}{K}$, where $K$ represents the number of identities.

For evaluation, we utilize a baseline VI-ReID model, namely AGW \cite{ye2021deep}, with noisy label smoothing. The results are presented in Table \ref{tab:models}. When compared to the best-performing GAN model, CUT, our VI-Diff model showcases enhancements of $2.45\%$ and $1.16\%$ in terms of All-search R1 and mAP, respectively, along with improvements of $3.61\%$ and $3.14\%$ in the Indoor-search scenario. On the other hand, in the case of diffusion models, both ILVR and EGSDE exhibit inferior results for this task. Specifically, ILVR demonstrates only marginal improvement over unconditional random generation. In contrast, EGSDE showcases a substantial performance gap of more than $15\%$ in R1 and over $12\%$ in mAP when compared to our VI-Diff model. This can be attributed to the low-pass filtered sampling guidance employed in these models, which introduces color information and leads to a loss of detail.

\subsection{Qualitative Evaluation}
\subsubsection{Translation Results Comparison}
\begin{figure}
    \centering
    \includegraphics[width=0.75\linewidth]{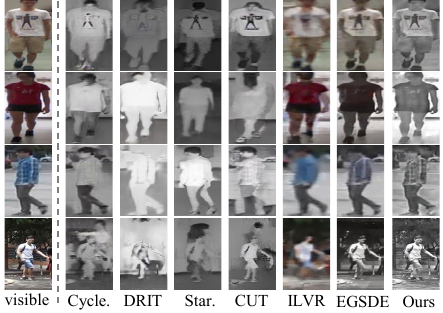}
    \caption{Cross-modality image translation (from visible to infrared). Cycle: CycleGAN; Star: StarGAN v2.}
    \label{fig:trans_sample}
\end{figure}

We compare VI-Diff with other translation models used in the experiments, and the qualitative results are presented in Fig. \ref{fig:trans_sample}. Our observations indicate that GAN methods tend to lose finer details during the translation process, resulting in influenced and distorted outcomes. Moreover, we observe that the two diffusion models (ILVR and EGSDE) are adversely affected by the color information introduced through low-pass filtered guidance. These qualitative findings are consistent with the quantitative comparisons depicted in Table \ref{tab:models}, further underscoring that VI-Diff attains superior visual quality and retains intricate details in comparison to these alternative methods.

\subsubsection{Visualization of Modality Gap in Conditions} 
\begin{figure}
    \centering
    \begin{subfigure}{0.4\linewidth}
        \includegraphics[width=1\linewidth]{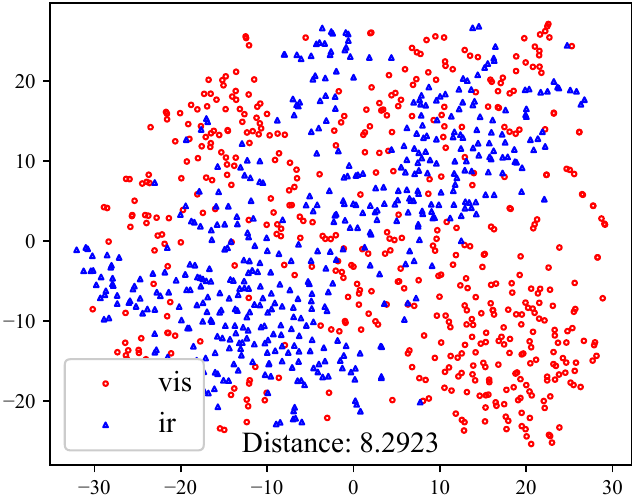}
        \caption{High-Pass Filtered}
        \label{tsne-high}
    \end{subfigure}
    \begin{subfigure}{0.4\linewidth}
        \includegraphics[width=1\linewidth]{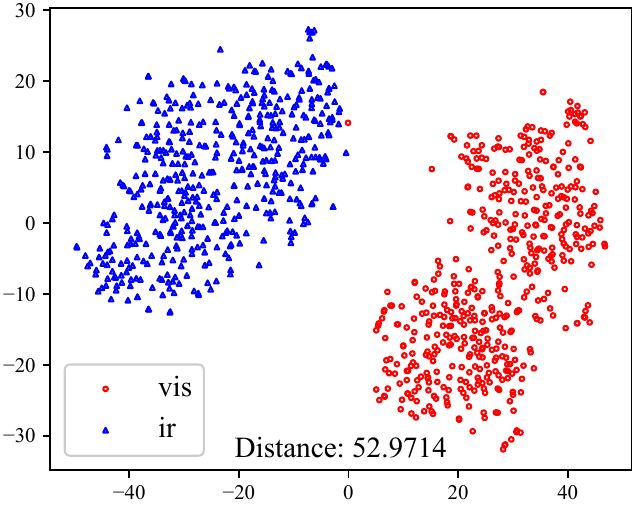}
        \caption{Low-Pass Filtered}
        \label{tsne-low}
    \end{subfigure}
    \caption{t-SNE visualization of modality gap in our high-pass filtered condition images and low-pass filtered images in ILVR/EGSDE. The distance is calculated using the centers of visualized features for each modality.}
    \label{fig:tsne}
\end{figure}

We employ t-SNE \cite{van2014accelerating} to visualize and compare the conditions (high-pass filtered images) used in our VI-Diff with the guidance (low-pass filtered images) used in the ILVR and EGSDE diffusion models, as depicted in Fig. \ref{fig:tsne}. To create these visualizations, we randomly select 500 visible and 500 infrared images. Subsequently, our high-pass filter and the low-pass filter from ILVR/EGSDE are individually applied to these images. To extract features from these processed images, we utilize a pre-trained ResNet50 \cite{he2016deep} model.

From Fig. \ref{fig:tsne}, it becomes evident that the high-pass filtered condition images from both modalities exhibit considerable overlap, suggesting a diminished modality gap. Conversely, the low-pass filtered images exhibit a pronounced modality gap. These visualization outcomes substantiate the assertions made in Fig. \ref{fig:highpass}, wherein the condition image in our VI-Diff is demonstrated to be modality-insensitive, while the low-pass filtered images in existing diffusion models (i.e., ILVR and EGSDE) exhibit a substantial modality gap.

\subsection{Ablation Study}
\subsubsection{Translation Without Condition $c$}
In order to showcase the efficacy of the condition $c=\mathcal{F}(x_0)$ in controlling content during translation, we conduct a comparison with a different content control technique that does not utilize $c$. This technique involves initiating denoising from a noised source image with an appropriate level of noise, as proposed by the diffusion model SDEdit \cite{meng2022sdedit}. The introduced noise conceals artifacts within the source image, while still retaining the overall structure, thus effectively employing the denoising process to generate an image with a structure akin to the source image.

\begin{figure}
    \centering
    \includegraphics[width=0.6\linewidth]{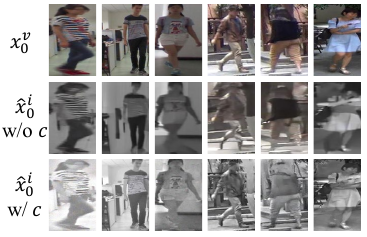}
    \caption{Translation without/with our condition $c$.}
    \label{fig:no_con_sample}
\end{figure}

We omit the condition $c$ from our VI-Diff model and train it with the same parameters using visible and infrared images, while utilizing the modality indicators ($\mathcal{D} = \{ \left(x_0^v, e^v \right), \left(x_0^i, e^i \right)\}$). During the translation process, the original initial noise $x_T$ is substituted with a noised source image at timestep $\frac{T}{2}$, i.e., $x_{T/2}^v$. The resulting translation outcomes are showcased in Fig. \ref{fig:no_con_sample}, clearly demonstrating that our VI-Diff with the condition $c$ is more adept at preserving intricate details.

\subsubsection{Effectiveness of Noisy Label Smoothing}
We conduct an evaluation to quantify the extent of improvement achievable through the application of GCE loss and LSR to address the noisy label issue, as discussed in \textit{Noisy Label Smoothing} section. We adapted the loss function in the AGW model and present the outcomes in Table \ref{tab:gcelsr}.

As depicted in Table \ref{tab:gcelsr}, both GCE and LSR result in improved performance of the AGW model within the single-modality labeled setting. GCE loss leads to enhancements of $1.13\%$ and $0.89\%$ in terms of R1 accuracy and mAP, respectively, in the All-search scenario. LSR loss produces even more significant improvements, elevating R1 and mAP in All-search by $2.26\%$ and $1.38\%$, respectively, and in Indoor-search by $1.2\%$ and $1.53\%$. However, combining GCE loss and LSR is less effective, potentially due to the differing nature of these regularization strategies. This experimentation underscores the efficacy of designing noisy label smoothing for generated cross-modality images, and it points towards the potential for further tailored loss strategies.

\begin{table}
\small
\centering
\begin{tabular}{llcccc}
\hline
{GCE} & {LSR} & \multicolumn{2}{c}{All-search}  & \multicolumn{2}{c}{Indoor-Search} \\ \cline{3-6} 
                     &                      & R1             & mAP            & R1              & mAP             \\ \hline
$\times$             & $\times$             & 22.43          & 23.59          & 29.68           & 39.12           \\
$\checkmark$         & $\times$             & 23.56          & 24.48          & 29.32           & 39.30           \\
$\times$             & $\checkmark$         & \textbf{24.69} & \textbf{24.97} & \textbf{30.88}  & \textbf{40.54}  \\
$\checkmark$         & $\checkmark$         & 22.93          & 23.83          & 29.33           & 38.97           \\ \hline
\end{tabular}
\caption{Ablation study of GCE loss and LSR.}
\label{tab:gcelsr}
\end{table}

\subsubsection{The percentage of overlapped IDs in training VI-Diff} In real-world scenarios, the collected VI-ReID dataset might not have matched ID from both modalities. We investigate under this situation how the percentage of overlapped IDs affects VI-Diff, as shown in Tab.\ref{tab:idpercent}. We randomly discard a number of different IDs in visible and infrared modality, until the remaining IDs in two sets has the desired percentage of overlap and then used to train VI-Diff. In VI-ReID model training, we translate all visible images to infrared for fair comparison. AGW is used in this experiment.

The results from Tab.\ref{tab:idpercent} shows that, imperfect matching ID between labeled visible and unlabeled infrared modality has not large influence on VI-Diff results. The performance difference between $0\%$ overlapped ID and $100\%$ overlapped ID is $1.28\%, 0.66\%$ in R1, mAP of All-search, and $0.52\%,0.56\%$ in R1, mAP of Indoor-search. Note that, after removing IDs, the training set of VI-Diff also shrinks, which could result in the performance drop.

\begin{table}
\small
\centering
\begin{tabular}{ccccc}
\hline
{Overlapped ID} & \multicolumn{2}{c}{All-search} & \multicolumn{2}{c}{Indoor-Search} \\ \cline{2-5} 
                            & R1             & mAP           & R1              & mAP             \\ \hline
0\%                           & 23.41          & 24.31         & 30.36           & 39.98           \\
50\%                          & 23.60          & 24.89         & 29.89           & 39.91           \\
100\%                         & 24.69          & 24.97         & 30.88           & 40.54           \\ \hline
\end{tabular}
\caption{Comparison on the percentage of overlapped ID of visible and infrared modality in training set of VI-Diff.}
\label{tab:idpercent}
\end{table}

\subsection{Comparison of Different VI-ReID models}
Compared to using the baseline VI-ReID model AGW, we conduct further evaluations using other state-of-the-art VI-ReID models to determine the potential performance gains within our single-modality labeled setting. The results are presented in Table \ref{tab:vireidmodel}. In this study, we select the DDAG\cite{ye2020dynamic}, CAJ\cite{ye2021channel}, DART\cite{Yang2022DART}, and DEEN\cite{zhang2023diverse} models, all of which have officially released code for reproducibility. Among these, CAJ and DART outperform the baseline AGW model. Specifically, CAJ demonstrates improvements of $3.58\%$ and $3.18\%$ in All-search R1 and mAP, respectively, along with enhancements of $3.7\%$ and $3.56\%$ in Indoor-search. This experimentation underscores the compatibility of our VI-Diff with various VI-ReID models, highlighting that adopting appropriate models can lead to further enhancements in task performance.

\begin{table}
\small
\centering
\begin{tabular}{llcccc}
\hline
{VI-ReID} & {Venue} & \multicolumn{2}{c}{All-search}  & \multicolumn{2}{c}{Indoor-Search} \\ \cline{3-6} 
                         &                        & R1             & mAP            & R1              & mAP             \\ \hline
AGW                      & \textit{TPAMI21}                & 24.69          & 24.97          & 30.88           & 40.54           \\
DDAG                     & \textit{ECCV20}                 & 23.10          & 24.13          & 29.63           & 39.32           \\
CAJ                      & \textit{ICCV21}                 & \textbf{28.27} & \textbf{28.15} & \textbf{34.58}  & 44.10           \\
DART                     & \textit{CVPR22}                 & 26.17          & 27.47          & 34.23           & \textbf{44.23}  \\
DEEN                     & \textit{CVPR23}                 & 23.15          & 23.73          & 28.75           & 38.87           \\ \hline
\end{tabular}
\caption{Comparison of different VI-ReID models.}
\label{tab:vireidmodel}
\end{table}

\subsection{Infrared to Visible Translation}
While visible images are more readily labeled due to their distinct characteristics, the practicality of the VI-ReID setting with only visible labels is evident. However, for the sake of comprehensiveness, we also conduct an experiment within an infrared-labeled setting, and the results are compared in Table \ref{tab:i2v}. Given the absence of color information in infrared images, the translated visible images of the same individual exhibit variations in clothing color, introducing a new challenge for VI-ReID models. As a result, a performance drop is observed in this experiment.

\begin{table}
\centering
\begin{tabular}{ccccc}
\hline
{translation} & \multicolumn{2}{c}{All-search} & \multicolumn{2}{c}{Indoor-Search} \\ \cline{2-5} 
                             & R1             & mAP           & R1              & mAP             \\ \hline
visible$\rightarrow$infrared                       & 24.69          & 24.97         & 30.88           & 40.54           \\
infrared$\rightarrow$visible                       & 15.55          & 15.75         & 23.74           & 32.15           \\ \hline
\end{tabular}
\caption{Compare infrared to visible translation. The AGW model is used.}
\label{tab:i2v}
\end{table}

%%%%%%%%%%%%%%%%%%%%%%%%%%%%%%%%%%%%%%%%%%%%%%%%%%%%%%%%%%%%%%%%%%%%%%%%%%%
\subsection{Experiments on RegDB}
\textbf{Dataset:}
The RegDB \cite{nguyen2017person} dataset comprises 412 different individuals, each with 10 visible images and 10 infrared images. We conduct ten experimental trials, each involving a distinct split of 206 identities for training and 206 identities for testing. These trials are designed to ensure consistent and stable results.

\textbf{Experiment Setting:}
The images in the RegDB dataset have a resolution of $64\times 128$, and they tend to be blurrier compared to those in the SYSU-MM01 dataset. To handle this difference in image quality, we use HED edge detection~\cite{xie15hed} to establish a modality-insensitive ID-relevant condition in our VI-Diff method during RegDB training. This choice is based on the reason: HED edge detection outperforms high-pass filtering in extracting contour information from RegDB dataset. HED can effectively understand and recognize complex image structures, enabling accurate edge extraction and reducing interference from noise and unnecessary details~\cite{xie15hed}.

\begin{figure}
    \centering
    \includegraphics[width=0.46\textwidth]{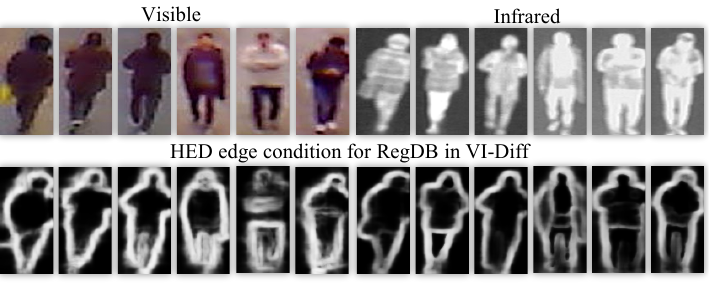}
    \caption{HED edge samples from RegDB.}
    \label{fig:hed}
\end{figure}

Consequently, for VI-Diff training on the RegDB dataset, we replace the high-pass filtered generation condition used for SYSU-MM01 with HED edge detection. This substitution is illustrated in Fig. \ref{fig:hed}. Other settings for RegDB remain the same as those for SYSU-MM01.

\textbf{Comparison with GANs And Diffusion Models:}
In the context of the RegDB dataset, we compare several models, including the most successful GAN model from SYSU-MM01, namely CUT \cite{Park2020CUT}. Additionally, we compare two diffusion models, ILVR \cite{Choi2021ILVRCM} and EGSDE \cite{zhao2022egsde}. The evaluation of the visible-to-infrared translation results is conducted using the benchmark VI-ReID model AGW \cite{ye2021deep}. The experimental results are presented in Table \ref{tab:regdb}.

When compared to CUT, our VI-Diff approach demonstrates enhancements of 5.4\% and 6.12\% in terms of R1 and mAP respectively, in the v2i search. Furthermore, in the i2v search, our VI-Diff model showcases improvements of 2.28\% and 2.97\% in R1 and mAP. It is important to note that the other two diffusion models continue to exhibit less favorable results on RegDB.

\begin{table}
\small
\centering
\begin{tabular}{lllll}
\hline
{Model} & \multicolumn{2}{l}{v2i search}  & \multicolumn{2}{l}{i2v search}  \\ \cline{2-5} 
        & R1             & mAP            & R1             & mAP            \\ \hline
CUT     & 26.36          & 23.48          & 23.80          & 21.55          \\ \hline
ILVR    & 2.34           &  3.32          & 1.09           & 2.33           \\
EGSDE   & 7.53           & 8.07           & 1.90           & 3.43            \\ \hline
VI-Diff & \textbf{31.76} & \textbf{29.60} & \textbf{26.08} & \textbf{24.52} \\ \hline
\end{tabular}
\caption{Comparison with CUT, ILVR, EGSDE. Results from AGW. (v2i search: visible to infrared search; i2v search: infrared to visible search.)}
\label{tab:regdb}
\end{table}

\textbf{Infrared to Visible Translation:}
We also assess the infrared-to-visible translation using the benchmark model AGW, as outlined in Table \ref{tab:regi2v}. Our observations indicate a decrease in performance in the visible-to-infrared (v2i) search scenario, whereas an improvement is evident in the infrared-to-visible (i2v) search. We posit that the issue of color loss during translation is not as pronounced as observed in SYSU-MM01. In the RegDB dataset, images of the same person in both visible and infrared are taken using synchronized RGB/IR cameras, leading to slight variations in posture. This characteristic bestows pose with a significant role in VI-ReID on the RegDB dataset. Furthermore, during the translation process of VI-Diff, the the body structure of the person is effectively preserved, thereby mitigating the decline in performance.

\begin{table}
\small
\centering
\begin{tabular}{ccccc}
\hline
{translation} & \multicolumn{2}{c}{v2i search} & \multicolumn{2}{c}{i2v search} \\ \cline{2-5} 
                             & R1             & mAP           & R1              & mAP             \\ \hline
visible$\rightarrow$infrared  & 31.76 & 29.60 & 26.08 & 24.52            \\
infrared$\rightarrow$visible  & 26.98 & 25.20 & 30.50 & 28.33 \\ \hline
\end{tabular}
\caption{Compare infrared to visible translation on RegDB. The AGW model is used.}
\label{tab:regi2v}
\end{table}

\textbf{Comparison of Different VI-ReID models:}
We employ the most successful VI-ReID model from our previous experiments (namely, CAJ~\cite{ye2021channel}) to assess the outcomes of visible-to-infrared translation using the data generated by our VI-Diff approach. The comparison is presented in Table \ref{tab:caj}. Notably, we observe improvements of 6.15\% and 4.09\% in terms of R1 and mAP respectively in the v2i search, along with gains of 3.19\% and 2.41\% in the R1 and mAP metrics in the i2v search. These results underscore the potential of leveraging cutting-edge VI-ReID models to enhance performance in this specific task.

\begin{table}
\small
\centering
\begin{tabular}{lllll}
\hline
{VI-ReID}  & \multicolumn{2}{l}{v2i search} & \multicolumn{2}{l}{i2v search} \\ \cline{2-5} 
                     & R1            & mAP            & R1            & mAP            \\ \hline
VI-Diff \textit{w.} AGW             & 31.76 & 29.60 & 26.08 & 24.52      \\
VI-Diff \textit{w.} CAJ             & 37.91 & 33.69 & 29.27 & 26.93 \\ \hline
\end{tabular}
\caption{Results on state-of-the-art VI-ReID model (CAJ). `w.' is short for `with'.}
\label{tab:caj}
\end{table}
\section{Conclusion}
In this study, we have delved deeply into the challenges posed by the single-modality labeled VI-ReID problem by focusing on the intricate realm of modality translation. Our novel solution, the VI-Diff model, introduced alongside a meticulously tailored condition framework, has exhibited remarkable efficacy in outperforming several established GANs and recent diffusion models. Our VI-Diff model not only attains state-of-the-art performance but also underscores the critical importance of addressing modality discrepancies in this context, ensuring that translated images retain essential identity attributes. In essence, this study represents a significant stride in employing unpaired image translation methods to effectively tackle the nuanced challenges of single-modality labeled VI-ReID.
{
    \small
    \bibliographystyle{ieeenat_fullname}
    \bibliography{main}

\begin{thebibliography}{42}
\providecommand{\natexlab}[1]{#1}
\providecommand{\url}[1]{\texttt{#1}}
\expandafter\ifx\csname urlstyle\endcsname\relax
  \providecommand{\doi}[1]{doi: #1}\else
  \providecommand{\doi}{doi: \begingroup \urlstyle{rm}\Url}\fi

\bibitem[Benaim and Wolf(2017)]{benaim2017one}
Sagie Benaim and Lior Wolf.
\newblock One-sided unsupervised domain mapping.
\newblock \emph{Advances in neural information processing systems}, 30, 2017.

\bibitem[Choi et~al.(2021)Choi, Kim, Jeong, Gwon, and Yoon]{Choi2021ILVRCM}
Jooyoung Choi, Sungwon Kim, Yonghyun Jeong, Youngjune Gwon, and Sungroh Yoon.
\newblock Ilvr: Conditioning method for denoising diffusion probabilistic
  models.
\newblock \emph{2021 IEEE/CVF International Conference on Computer Vision
  (ICCV)}, pages 14347--14356, 2021.

\bibitem[Choi et~al.(2020{\natexlab{a}})Choi, Lee, Kim, Kim, and
  Kim]{choi2020hi}
Seokeon Choi, Sumin Lee, Youngeun Kim, Taekyung Kim, and Changick Kim.
\newblock Hi-cmd: Hierarchical cross-modality disentanglement for
  visible-infrared person re-identification.
\newblock In \emph{Proceedings of the IEEE/CVF conference on computer vision
  and pattern recognition}, pages 10257--10266, 2020{\natexlab{a}}.

\bibitem[Choi et~al.(2020{\natexlab{b}})Choi, Uh, Yoo, and Ha]{choi2020stargan}
Yunjey Choi, Youngjung Uh, Jaejun Yoo, and Jung-Woo Ha.
\newblock Stargan v2: Diverse image synthesis for multiple domains.
\newblock In \emph{Proceedings of the IEEE/CVF conference on computer vision
  and pattern recognition}, pages 8188--8197, 2020{\natexlab{b}}.

\bibitem[Dai et~al.(2018)Dai, Ji, Wang, Wu, and Huang]{dai2018cross}
Pingyang Dai, Rongrong Ji, Haibin Wang, Qiong Wu, and Yuyu Huang.
\newblock Cross-modality person re-identification with generative adversarial
  training.
\newblock In \emph{Proceedings of the 27th International Joint Conference on
  Artificial Intelligence}, page 677–683. AAAI Press, 2018.

\bibitem[Deng et~al.(2018)Deng, Zheng, Ye, Kang, Yang, and Jiao]{deng2018image}
Weijian Deng, Liang Zheng, Qixiang Ye, Guoliang Kang, Yi Yang, and Jianbin
  Jiao.
\newblock Image-image domain adaptation with preserved self-similarity and
  domain-dissimilarity for person re-identification.
\newblock In \emph{Proceedings of the IEEE conference on computer vision and
  pattern recognition}, pages 994--1003, 2018.

\bibitem[Dhariwal and Nichol(2021)]{dhariwal2021diffusion}
Prafulla Dhariwal and Alexander Nichol.
\newblock Diffusion models beat gans on image synthesis.
\newblock \emph{Advances in neural information processing systems},
  34:\penalty0 8780--8794, 2021.

\bibitem[Fu et~al.(2022)Fu, Huang, Zhou, Ma, Xu, and Zhang]{Fu2022c2}
Xiaowei Fu, Fuxiang Huang, Yuhang Zhou, Huimin Ma, Xin Xu, and Lei Zhang.
\newblock Cross-modal cross-domain dual alignment network for rgb-infrared
  person re-identification.
\newblock \emph{IEEE Transactions on Circuits and Systems for Video
  Technology}, 32\penalty0 (10):\penalty0 6874--6887, 2022.

\bibitem[He et~al.(2016)He, Zhang, Ren, and Sun]{he2016deep}
Kaiming He, Xiangyu Zhang, Shaoqing Ren, and Jian Sun.
\newblock Deep residual learning for image recognition.
\newblock In \emph{Proceedings of the IEEE conference on computer vision and
  pattern recognition}, pages 770--778, 2016.

\bibitem[Ho and Salimans(2022)]{ho2022classifier}
Jonathan Ho and Tim Salimans.
\newblock Classifier-free diffusion guidance.
\newblock \emph{arXiv preprint arXiv:2207.12598}, 2022.

\bibitem[Ho et~al.(2020)Ho, Jain, and Abbeel]{ho2020denoising}
Jonathan Ho, Ajay Jain, and Pieter Abbeel.
\newblock Denoising diffusion probabilistic models.
\newblock \emph{Advances in neural information processing systems},
  33:\penalty0 6840--6851, 2020.

\bibitem[Huang et~al.(2019)Huang, Wu, Xu, and Zhong]{huang2019sbsgan}
Yan Huang, Qiang Wu, JingSong Xu, and Yi Zhong.
\newblock Sbsgan: Suppression of inter-domain background shift for person
  re-identification.
\newblock In \emph{Proceedings of the IEEE/CVF International Conference on
  Computer Vision}, pages 9527--9536, 2019.

\bibitem[Kim et~al.(2017)Kim, Cha, Kim, Lee, and Kim]{kim2017learning}
Taeksoo Kim, Moonsu Cha, Hyunsoo Kim, Jung~Kwon Lee, and Jiwon Kim.
\newblock Learning to discover cross-domain relations with generative
  adversarial networks.
\newblock In \emph{International conference on machine learning}, pages
  1857--1865. PMLR, 2017.

\bibitem[Lee et~al.(2018)Lee, Tseng, Huang, Singh, and Yang]{DRIT}
Hsin-Ying Lee, Hung-Yu Tseng, Jia-Bin Huang, Maneesh~Kumar Singh, and
  Ming-Hsuan Yang.
\newblock Diverse image-to-image translation via disentangled representations.
\newblock In \emph{European Conference on Computer Vision}, 2018.

\bibitem[Liang et~al.(2021{\natexlab{a}})Liang, Wang, Lai, and Xie]{LiangH2H}
Wenqi Liang, Guangcong Wang, Jianhuang Lai, and Xiaohua Xie.
\newblock Homogeneous-to-heterogeneous: Unsupervised learning for rgb-infrared
  person re-identification.
\newblock \emph{IEEE Transactions on Image Processing}, 30:\penalty0
  6392--6407, 2021{\natexlab{a}}.

\bibitem[Liang et~al.(2021{\natexlab{b}})Liang, Wang, Lai, and
  Xie]{liang2021homogeneous}
Wenqi Liang, Guangcong Wang, Jianhuang Lai, and Xiaohua Xie.
\newblock Homogeneous-to-heterogeneous: Unsupervised learning for rgb-infrared
  person re-identification.
\newblock \emph{IEEE Transactions on Image Processing}, 30:\penalty0
  6392--6407, 2021{\natexlab{b}}.

\bibitem[Lu et~al.(2020)Lu, Wu, Liu, Zhang, Li, Chu, and Yu]{lu2020cross}
Yan Lu, Yue Wu, Bin Liu, Tianzhu Zhang, Baopu Li, Qi Chu, and Nenghai Yu.
\newblock Cross-modality person re-identification with shared-specific feature
  transfer.
\newblock In \emph{Proceedings of the IEEE/CVF Conference on Computer Vision
  and Pattern Recognition}, pages 13379--13389, 2020.

\bibitem[Meng et~al.(2022)Meng, He, Song, Song, Wu, Zhu, and
  Ermon]{meng2022sdedit}
Chenlin Meng, Yutong He, Yang Song, Jiaming Song, Jiajun Wu, Jun-Yan Zhu, and
  Stefano Ermon.
\newblock {SDE}dit: Guided image synthesis and editing with stochastic
  differential equations.
\newblock In \emph{International Conference on Learning Representations}, 2022.

\bibitem[Nguyen et~al.(2017)Nguyen, Hong, Kim, and Park]{nguyen2017person}
Dat~Tien Nguyen, Hyung~Gil Hong, Ki~Wan Kim, and Kang~Ryoung Park.
\newblock Person recognition system based on a combination of body images from
  visible light and thermal cameras.
\newblock \emph{Sensors}, 17\penalty0 (3):\penalty0 605, 2017.

\bibitem[Park et~al.(2021)Park, Lee, Lee, and Ham]{park2021learning}
Hyunjong Park, Sanghoon Lee, Junghyup Lee, and Bumsub Ham.
\newblock Learning by aligning: Visible-infrared person re-identification using
  cross-modal correspondences.
\newblock In \emph{Proceedings of the IEEE/CVF International Conference on
  Computer Vision}, pages 12046--12055, 2021.

\bibitem[Park et~al.(2020)Park, Efros, Zhang, and Zhu]{Park2020CUT}
Taesung Park, Alexei~A. Efros, Richard Zhang, and Jun-Yan Zhu.
\newblock Contrastive learning for unpaired image-to-image translation.
\newblock In \emph{Computer Vision -- ECCV 2020}, pages 319--345, Cham, 2020.
  Springer International Publishing.

\bibitem[Ronneberger et~al.(2015)Ronneberger, Fischer, and
  Brox]{ronneberger2015u}
Olaf Ronneberger, Philipp Fischer, and Thomas Brox.
\newblock U-net: Convolutional networks for biomedical image segmentation.
\newblock In \emph{Medical Image Computing and Computer-Assisted
  Intervention--MICCAI 2015: 18th International Conference, Munich, Germany,
  October 5-9, 2015, Proceedings, Part III 18}, pages 234--241. Springer, 2015.

\bibitem[Saharia et~al.(2022)Saharia, Chan, Chang, Lee, Ho, Salimans, Fleet,
  and Norouzi]{saharia2022palette}
Chitwan Saharia, William Chan, Huiwen Chang, Chris Lee, Jonathan Ho, Tim
  Salimans, David Fleet, and Mohammad Norouzi.
\newblock Palette: Image-to-image diffusion models.
\newblock In \emph{ACM SIGGRAPH 2022 Conference Proceedings}, pages 1--10,
  2022.

\bibitem[Song et~al.(2020)Song, Meng, and Ermon]{song2020denoising}
Jiaming Song, Chenlin Meng, and Stefano Ermon.
\newblock Denoising diffusion implicit models.
\newblock \emph{arXiv preprint arXiv:2010.02502}, 2020.

\bibitem[Su et~al.(2023)Su, Song, Meng, and Ermon]{su2022dual}
Xuan Su, Jiaming Song, Chenlin Meng, and Stefano Ermon.
\newblock Dual diffusion implicit bridges for image-to-image translation.
\newblock In \emph{International Conference on Learning Representations}, 2023.

\bibitem[Szegedy et~al.(2016)Szegedy, Vanhoucke, Ioffe, Shlens, and
  Wojna]{szegedy2016rethinking}
Christian Szegedy, Vincent Vanhoucke, Sergey Ioffe, Jon Shlens, and Zbigniew
  Wojna.
\newblock Rethinking the inception architecture for computer vision.
\newblock In \emph{Proceedings of the IEEE conference on computer vision and
  pattern recognition}, pages 2818--2826, 2016.

\bibitem[Van Der~Maaten(2014)]{van2014accelerating}
Laurens Van Der~Maaten.
\newblock Accelerating t-sne using tree-based algorithms.
\newblock \emph{The journal of machine learning research}, 15\penalty0
  (1):\penalty0 3221--3245, 2014.

\bibitem[Wang et~al.(2022)Wang, Zhang, Chen, Zhang, Wang, Sheng, Qu, and
  Xie]{wang2022optimal}
Jiangming Wang, Zhizhong Zhang, Mingang Chen, Yi Zhang, Cong Wang, Bin Sheng,
  Yanyun Qu, and Yuan Xie.
\newblock Optimal transport for label-efficient visible-infrared person
  re-identification.
\newblock In \emph{European Conference on Computer Vision}, pages 93--109.
  Springer, 2022.

\bibitem[Wei et~al.(2018)Wei, Zhang, Gao, and Tian]{wei2018person}
Longhui Wei, Shiliang Zhang, Wen Gao, and Qi Tian.
\newblock Person transfer gan to bridge domain gap for person
  re-identification.
\newblock In \emph{Proceedings of the IEEE conference on computer vision and
  pattern recognition}, pages 79--88, 2018.

\bibitem[Wu et~al.(2017)Wu, Zheng, Yu, Gong, and Lai]{wu2017rgb}
Ancong Wu, Wei-Shi Zheng, Hong-Xing Yu, Shaogang Gong, and Jianhuang Lai.
\newblock Rgb-infrared cross-modality person re-identification.
\newblock In \emph{Proceedings of the IEEE international conference on computer
  vision}, pages 5380--5389, 2017.

\bibitem[Wu et~al.(2021)Wu, Dai, Chen, Lin, Wu, Huang, Zhong, and
  Ji]{WU2021mpanet}
Qiong Wu, Pingyang Dai, Jie Chen, Chia-Wen Lin, Yongjian Wu, Feiyue Huang,
  Bineng Zhong, and Rongrong Ji.
\newblock Discover cross-modality nuances for visible-infrared person
  re-identification.
\newblock In \emph{2021 IEEE/CVF Conference on Computer Vision and Pattern
  Recognition (CVPR)}, pages 4328--4337, 2021.

\bibitem[Xie and Tu(2015)]{xie15hed}
Saining Xie and Zhuowen Tu.
\newblock Holistically-nested edge detection.
\newblock In \emph{Proceedings of IEEE International Conference on Computer
  Vision}, 2015.

\bibitem[Yang et~al.(2022)Yang, Huang, Hu, Li, Lv, and Peng]{Yang2022DART}
Mouxing Yang, Zhenyu Huang, Peng Hu, Taihao Li, Jiancheng Lv, and Xi Peng.
\newblock Learning with twin noisy labels for visible-infrared person
  re-identification.
\newblock In \emph{Proceedings of the IEEE/CVF Conference on Computer Vision
  and Pattern Recognition (CVPR)}, pages 14308--14317, 2022.

\bibitem[Ye et~al.(2019)Ye, Lan, Wang, and Yuen]{ye2019bi}
Mang Ye, Xiangyuan Lan, Zheng Wang, and Pong~C Yuen.
\newblock Bi-directional center-constrained top-ranking for visible thermal
  person re-identification.
\newblock \emph{IEEE Transactions on Information Forensics and Security},
  15:\penalty0 407--419, 2019.

\bibitem[Ye et~al.(2020)Ye, Shen, J~Crandall, Shao, and Luo]{ye2020dynamic}
Mang Ye, Jianbing Shen, David J~Crandall, Ling Shao, and Jiebo Luo.
\newblock Dynamic dual-attentive aggregation learning for visible-infrared
  person re-identification.
\newblock In \emph{European Conference on Computer Vision}, pages 229--247.
  Springer, 2020.

\bibitem[Ye et~al.(2021{\natexlab{a}})Ye, Ruan, Du, and Shou]{ye2021channel}
Mang Ye, Weijian Ruan, Bo Du, and Mike~Zheng Shou.
\newblock Channel augmented joint learning for visible-infrared recognition.
\newblock In \emph{Proceedings of the IEEE/CVF International Conference on
  Computer Vision}, pages 13567--13576, 2021{\natexlab{a}}.

\bibitem[Ye et~al.(2021{\natexlab{b}})Ye, Shen, Lin, Xiang, Shao, and
  Hoi]{ye2021deep}
Mang Ye, Jianbing Shen, Gaojie Lin, Tao Xiang, Ling Shao, and Steven~CH Hoi.
\newblock Deep learning for person re-identification: A survey and outlook.
\newblock \emph{IEEE transactions on pattern analysis and machine
  intelligence}, 44\penalty0 (6):\penalty0 2872--2893, 2021{\natexlab{b}}.

\bibitem[Yi et~al.(2017)Yi, Zhang, Tan, and Gong]{yi2017dualgan}
Zili Yi, Hao Zhang, Ping Tan, and Minglun Gong.
\newblock Dualgan: Unsupervised dual learning for image-to-image translation.
\newblock In \emph{Proceedings of the IEEE international conference on computer
  vision}, pages 2849--2857, 2017.

\bibitem[Zhang and Wang(2023)]{zhang2023diverse}
Yukang Zhang and Hanzi Wang.
\newblock Diverse embedding expansion network and low-light cross-modality
  benchmark for visible-infrared person re-identification.
\newblock In \emph{Proceedings of the IEEE/CVF Conference on Computer Vision
  and Pattern Recognition}, pages 2153--2162, 2023.

\bibitem[Zhang and Sabuncu(2018)]{zhang2018generalized}
Zhilu Zhang and Mert Sabuncu.
\newblock Generalized cross entropy loss for training deep neural networks with
  noisy labels.
\newblock \emph{Advances in neural information processing systems}, 31, 2018.

\bibitem[Zhao et~al.(2022)Zhao, Bao, Li, and Zhu]{zhao2022egsde}
Min Zhao, Fan Bao, Chongxuan Li, and Jun Zhu.
\newblock Egsde: Unpaired image-to-image translation via energy-guided
  stochastic differential equations.
\newblock \emph{Advances in Neural Information Processing Systems},
  35:\penalty0 3609--3623, 2022.

\bibitem[Zhu et~al.(2017)Zhu, Park, Isola, and Efros]{zhu2017unpaired}
Jun-Yan Zhu, Taesung Park, Phillip Isola, and Alexei~A Efros.
\newblock Unpaired image-to-image translation using cycle-consistent
  adversarial networks.
\newblock In \emph{Proceedings of the IEEE international conference on computer
  vision}, pages 2223--2232, 2017.

\end{thebibliography}
}


\end{document}